\documentclass{article}
\usepackage{kr}

\usepackage{times}
\usepackage{soul}
\usepackage{url}
\usepackage[hidelinks]{hyperref}
\usepackage[utf8]{inputenc}
\usepackage[small]{caption}
\usepackage{graphicx}
\usepackage{amsmath}
\usepackage{amsthm}
\usepackage{booktabs}
\usepackage{algorithm}
\usepackage{algorithmic}
\usepackage{amsfonts}%
\usepackage{bm}%
\usepackage{array}%
\usepackage{multirow}%
\usepackage{float}%
\usepackage{natbib}%

\title{Dynamic Domain Adaptation-Driven Physics-Informed Graph \\ Representation Learning for AC-OPF}

\author{Hongjie Zhu$^1$~~ Zezheng Zhang$^2$~~ Zeyu Zhang$^3$~~ Yu Bai$^1$~~ Shimin Wen$^1$~~ Huazhang Wang$^4$\\ 
Daji Ergu$^1$~~ Ying Cai$^1$\thanks{Corresponding author: caiying34@yeah.net}~~ Yang Zhao$^3$\\\vspace{0.2cm}
$^1$Southwest Minzu University~~
$^2$Northeast Electric Power University\\
$^3$La Trobe University~~
$^4$Institute of Tibetan Plateau Research
}

\begin{document}

\maketitle

\begin{abstract}
Alternating Current Optimal Power Flow (AC-OPF) seeks to optimize generator power injections by leveraging the non-linear relationships between voltage magnitudes and phase angles within the power system. However, current AC-OPF solvers face challenges in effectively capturing and representing the complex mapping between distributions of variables in the constraint space and their corresponding optimal solution variables. Consequently, this singular constraint mechanism restricts the network's capacity to develop a diverse knowledge representation space. Moreover, modeling the intricate relationships within the power grid solely based on spatial topology additionally constrains the network's capacity to integrate further prior knowledge (e.g., temporal information). To address these challenges, we propose \textbf{DDA-PIGCN}, \textbf{D}ynamic \textbf{D}omain \textbf{A}daptation-Driven \textbf{P}hysics-\textbf{I}nformed \textbf{G}raph \textbf{C}onvolutional \textbf{N}etwork (GCN), a novel approach designed to handle constraint-related challenges and develop a graph knowledge representation learning strategy that integrates spatiotemporal features. Specifically, DDA-PIGCN enhances consistency optimization for features with varying long-range dependencies by incorporating multi-layer, hard physics-informed constraints. Furthermore, by employing a learning approach based on the dynamic domain adaptation mechanism, and iteratively updating and feeding back key state variables under predefined constraints, DDA-PIGCN offers precise support for subsequent constraint verification. It also models and captures the spatiotemporal dependencies between generators and loads, fully utilizing the physical topology of the power grid, thereby achieving deep integration of topological data across both temporal and spatial dimensions. Comprehensive comparative and ablation experiments demonstrate that DDA-PIGCN achieves remarkable performance gains across multiple IEEE standard cases (case9, case30, and case300, etc.), with MAE ranging from \textbf{0.0011} to \textbf{0.0624} and constraint satisfaction rates between \textbf{99.6\%} and \textbf{100.0\%}, proving it to be a reliable and efficient AC-OPF solver.
\end{abstract}

\section{Introduction}

As a critical analytical framework for power system optimization, Alternating Current Optimal Power Flow (AC-OPF) computation serves as the computational cornerstone for modern grid operational decision-making. The objective is to calculate the optimal power injections for generators, satisfying each load's demand while strictly adhering to the physical constraints of key components and overall system operational limits. Traditionally, methods~\citep{baker2021solutions,kim2022projection,yang2024dc} that leverage the Direct Current (DC) approximation derive the linearized form of the power flow equations based on small-angle approximations. Additionally, methods~\citep{10406998,10050141} based on Second-Order Cone Programming (SOCP) can accurately model non-linear power flow constraints by incorporating convex relaxation techniques, providing a more precise and efficient optimization framework. However, the complexity of the solution space, the enormous computational workload, and the prevalence of local optima make it challenging for previous methods to effectively handle high-dimensional coupling, dynamic constraints, and uncertainties.

\begin{figure}[t]
    \centering
    \includegraphics[width=\columnwidth]{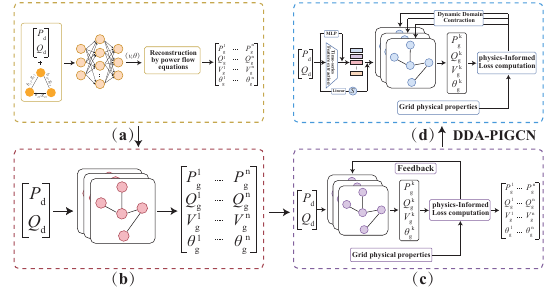}  
    \caption{(a)–(c) depict three previously proposed methods for solving the AC-OPF problem: traditional neural networks~\citep{han2024frmnet}, graph convolutional neural networks~\citep{gao2023physics}, and neural networks with physics-informed constraints~\citep{varbella2024physics}. (d) represents our proposed method.}
    \label{fig_1}
\end{figure}

The recent emergence of deep learning models has facilitated the integration of data-driven approaches with power systems, gradually establishing these models as essential tools for solving the AC-OPF problem. Some studies~\citep{huang2021deepopf,han2024frmnet} leverage the multi-layer non-linear modeling capabilities of deep neural networks (DNNs), training on large amounts of historical or simulated data (such as grid topology, loads, and generator parameters) to learn the input-output mapping of the AC-OPF problem and effectively approximate its optimal solution. Additionally, several works~\citep{gao2023physics,ghamizi2024opf} leverage graph convolutional networks (GCNs)~\citep{scarselli2008graph}, which model the power grid as a graph structure to simulate its topology and dynamic characteristics, capturing the complex relationships between buses (nodes) and lines (edges). Nevertheless, relying solely on complex mapping predictions makes it difficult to effectively represent topological knowledge because the network lacks effective strategies to satisfy AC-OPF equality and inequality constraints.

Physics-Informed Neural Networks (PINNs)\citep{raissi2019physics,lu2021physics} solve problems governed by partial differential equations (PDEs) by embedding physical constraints directly into neural networks, making them well-suited for handling equality and inequality constraints. PINNs offer a novel way to tackle nonlinear AC-OPF problems, overcoming challenges in complex knowledge representation. For instance, Nellikkath et al. (\citeyear{nellikkath2022physics}) integrate Karush-Kuhn-Tucker (KKT) conditions into the AC-OPF network, combining data-driven learning with physical constraints. Gao et al. (\citeyear{gao2023physics}) introduce PINNs indirectly as an additional supervisory signal by solving equations with predicted values as inputs. Figure~\ref{fig_1} shows the evolution of AC-OPF solution methods. However, these methods typically apply physical constraint losses only during feature decoupling, neglecting the crucial need to detect out-of-bound states throughout the entire process—from generation and load to final output. A lack of sufficient constraints can lead to abnormal updates and hinder effective knowledge representation learning, preventing optimal power generation. To address this, this paper proposes a dynamic domain adaptation-driven physics-informed method. It imposes dynamically varying constraints at different computation stages and adapts them based on feedback from each layer, enabling better long-range dependency modeling and more precise feature extraction within the PINNs framework.

In addition, the AC-OPF load forecasting problem requires comprehensive analysis from multiple physical perspectives in power systems. Traditional methods~\citep{gao2023physics,ghamizi2024opf} often rely on node features and model them using GNNs, but overlook the complex geometric relationships and coupling characteristics in graph-structured data. This is especially true in power systems, where the predicted values of different load nodes are closely related to the physical distance between the loads and generators. To address this issue, an improved module is proposed that integrates sequential graph-structured features. First, the graph theory algorithm is used to reorder the graph data based on spatial attributes and then serialize the data. Next, one-dimensional convolution operations are applied to extract meaningful features from the serialized data, serving as additional inputs to enrich the knowledge representation of the GNN model. This strategy aims to capture a more comprehensive network topology and latent temporal characteristics to accurately reflect the dynamic operation of the power system.

The main contributions of this paper are summarized as follows:
 
    \begin{itemize}
    
    \item We propose a novel framework, DDA-PIGCN, for solving the AC-OPF problem. DDA-PIGCN is a GCN that integrates multi-level physical constraints to guide the feature mapping of variables with varying long-range dependency characteristics using PINN. This approach aligns feature learning more closely with physical laws and system priors, offering a new paradigm for knowledge representation learning on graph-structured data.
    
    \item Unlike traditional static constraint strategies, DDA-PIGCN introduces a dynamic domain adaptation mechanism that incorporates feedback-driven physical constraints throughout the network’s layer-by-layer knowledge learning, effectively guiding the model to strictly adhere to the system’s normal operating range.
    
    \item We also design a temporal multi-feature extraction module as a complementary optimization to DDA-PIGNN. The spatial dependencies of non-Euclidean data and extractions of various temporal features are simulated by this module from a multi-feature perspective, with additional prior features being provided for the subsequent network.
 
    \end{itemize}

\section{Related Work}

\subsection{Alternating Current Optimal Power Flow}

The AC-OPF problem aims to minimize generation costs or power losses by optimizing generator outputs, voltage magnitudes, and power flow distribution, while maintaining the stable operation of the power system~\citep{gao2023physics,10050141}. The inputs include load demands, grid topology, and physical constraints (such as voltage limits, power bounds, and generator output capacities), whereas the outputs comprise optimized generator outputs, voltage profiles, and power flow distribution. During the solution process, several constraints must be satisfied, including power balance, voltage magnitude limits at nodes, generator output limits, and transmission line flow limits. Addressing the AC-OPF problem is crucial not only for ensuring the stable operation of power system components but also for minimizing resource waste and enhancing energy utilization efficiency. As electricity demand grows and grid topologies become more complex, efficient resource allocation and optimized scheduling have emerged as urgent challenges~\citep{xiang2019deepopf,10406998}.

\subsection{Conventional Methods}

To tackle this challenge, prior studies have utilized DC approximation~\citep{xiang2019deepopf} to emulate the small-angle approximation in AC power flow equations. Additionally, Lavaei et al. (\citeyear{lavaei2010convexification}) and Low et al. (\citeyear{low2014convex}) employ convex relaxation techniques in solvers to closely approximate the optimal solution state. Concurrently, second-order cone programming approximations have been thoroughly investigated to improve the accuracy of AC-OPF solution values~\citep{chowdhury2023real,chowdhury2023distributed}. However, converting AC-OPF into a convex optimization problem to derive an optimal solution overlooks the computational burden, as well as the timeliness and safety requirements of grid load forecasting~\citep{baker2021solutions,kile2014comparison}. Consequently, this method proves impractical for deployment on large-scale test cases, such as IEEE Case 118. Furthermore, these approaches frequently lack a comprehensive understanding of nonlinear equations and depend heavily on linear approximations, potentially yielding suboptimal solutions and introducing notable limitations. The proposed approach combines a Graph Convolutional Network (GCN)~\citep{scarselli2008graph} tailored to power grid topology with Physics-Informed Neural Networks (PINNs)~\citep{raissi2019physics,lu2021physics}, guiding the model to enforce predefined equality and inequality constraints, thereby optimizing solution accuracy and better aligning with real-world power grid operating conditions.

\subsection{Deep Learning-based Models}

In recent years, Pan et al. (\citeyear{pan2021deepopf}) propose leveraging deep neural networks (DNNs) to construct the mapping between load inputs and dispatch and power flow outputs, while employing a uniform sampling strategy to address the model’s overfitting issue. Jia et al. (\citeyear{jia2022convopf}) integrate cluster analysis into convolutional neural networks, utilizing historical network operating states to model the topological labels of location data. This method yields nearly a hundredfold increase in computational speed while maintaining high-accuracy solutions. At the same time, the advent of GNNs~\citep{scarselli2008graph} offers a novel research perspective for non-Euclidean data structures, such as power grid data. Numerous researchers have made substantial contributions to addressing the AC-OPF problem through the application of GNNs. For instance, Donon et al. (\citeyear{donon2020neural}) pioneer the application of GNN to optimize the AC-OPF problem, utilizing Kirchhoff’s law to impose constraints on the nodes. This approach significantly accelerates the solution process while enhancing computational efficiency. Piloto et al. (\citeyear{piloto2024canos}) demonstrate through extensive experiments that supervised GNNs can accurately predict AC-OPF solutions. By constructing a sophisticated mapping between load and power generation, they provide valuable insights into addressing the AC-OPF problem. However, relying solely on complex mappings to capture implicit knowledge in graph topology is often inadequate, since predicted values at load nodes closely depend on their physical distance to generators, which affects power transmission costs, and generators connected to multiple loads generally need to supply higher power.

\subsection{Physical-Informed Constraint}

Physical constraints are introduced by adding extra loss related to the physical problem, aiming to ensure that the method follows the correct physical laws during the learning process. Raissi et al. (\citeyear{raissi2019physics}) first introduce physics-informed neural networks to solve classical problems controlled by differential equations, such as fluid dynamics, quantum mechanics, and nonlinear shallow water wave propagation. Gao et al. (\citeyear{gao2023physics}) derive variables that satisfy physical ranges from the power flow equations in AC-OPF as additional supervision to train the GNN, and utilize an iterative feature construction method to encode physical features and practical constraints into the node vectors, thereby optimizing the accuracy of the solution. Varbella et al. (\citeyear{varbella2024physics}) use hard constraints (hPINNs)~\citep{lu2021physics} to guide the hierarchical GNN to satisfy specific boundaries or limitations, ensuring that each solution meets these conditions through mandatory hard constraints, thereby significantly improving the solution's effectiveness. However, most methods only apply physics-informed constraints in the decoder to guide network learning, neglecting the correlation between the target predictions and key features at different solving stages in the AC-OPF problem, thereby failing to fully leverage the advantages of the constraints. Therefore, we propose DDA-PIGCN, which applies dynamically adaptable physical constraints to each layer of the network, guiding the key features of each layer to support the final prediction.

\section{Method}

\subsection{Preliminaries of GCN}

Graph Convolutional Networks (GCNs) can effectively capture and process the topology and intricate relationships within the power grid, enabling deep modeling and knowledge representation of grid information. The GCN operates on graph-structured data represented as \( \mathcal{G} = (V, E) \), where \( V \in \mathbb{R}^{n \times m} \) denotes the node feature matrix, with \( n \) representing the total number of nodes and \( m \) the feature dimension of each node. Similarly, \( E \in \mathbb{R}^{b \times d} \) denotes the edge feature matrix, where \( b \) is the total number of edges and \( d \) is the feature dimension of each edge. The adjacency matrix \( A = \{ a_{i,j} \mid i, j = 1, 2, \ldots, n \} \in \mathbb{R}^{n \times n} \) encodes the connectivity relationships between nodes. Specifically, \( a_{ij} = 1 \) indicates the presence of an edge between nodes \(i\) and \(j\); otherwise, \( a_{ij} = 0 \). The \(A\) is typically normalized using the diagonal degree matrix \(D = [d_1, d_2, \ldots, d_n]_{\text{diag}}\), where \(d_i = \sum_{j=1}^n a_{ij}\) denotes the degree of the corresponding nodes \(i\). The primary normalization methods include random walk normalization (\ref{random walk normalization}), enhanced Laplacian normalization (\ref{enhanced Laplacian normalization}), and normalized Laplacian smoothing (\ref{normalized Laplacian smoothing}), with their mathematical formulations presented below:

    \begin{figure*}[t] 
    \centering
    \includegraphics[width=\textwidth]{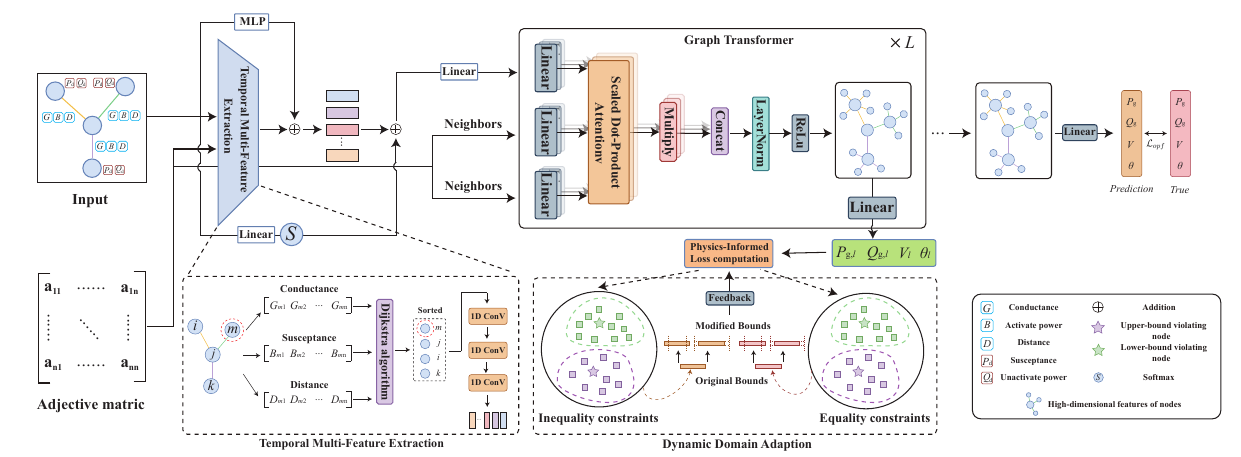}
    \caption{An overview of the proposed DDA-PIGCN. It comprises three pivotal components: Firstly, a Multi-Physics-Informed Constraint module is designed to guide the feature mapping of variables with different long-range dependency characteristics. Secondly, the Dynamic Domain Adaption mechanism adaptively adjusts the constraint conditions based on feedback obtained from each layer of the network. Lastly, a Temporal Multi-Feature Extraction module mines temporal features within the graph topology, supplying additional prior information to subsequent network layers.}
    \label{pipeline_fig} 
    \end{figure*}

\begin{equation}
    \hat{A} = D^{-1} A
\label{random walk normalization}
\end{equation}

\begin{equation}
    \hat{A} = I - D^{-\frac{1}{2}} A D^{-\frac{1}{2}}
\label{enhanced Laplacian normalization}
\end{equation}

\begin{equation}
    \hat{A} = D^{-\frac{1}{2}} (A + I) D^{-\frac{1}{2}}
\label{normalized Laplacian smoothing}
\end{equation}
 
\noindent where \(\hat{A}\) denotes the normalized adjacency matrix, and \(I\) represents the identity matrix. The GCN iteratively aggregates multisets of feature vectors from each node and its neighbors through a series of message-passing layers. The message-passing process can be described as follows:

\begin{equation}
    x_i^{(k+1)} = \text{comb}_{k+1} \left( x_v^{(k)}, \text{agg}_{k+1} \left( \{ x_u^{(k)} : u \in ne(i) \} \right) \right),
\end{equation}

\noindent where $0 \leq k \leq K-1 $, and \( x_{i}^{(k)} \) denotes the feature vector of node $i$ at the $k$-th iteration. The families of functions $\{\text{comb}_{k}\}_{k=1,...,K}$ and $\{\text{agg}_{k}\}_{k=1,...,K}$ are respectively defined for each iteration up to depth \(K\). Subsequently, the message-passing mechanism aggregates information from the neighbor set $ne(i)$ of node $i$ (i.e., nodes directly connected to node $i$) via the aggregation function $\text{agg}_{k}$. The aggregated neighbor information is combined with the original features of node $i$ by the combination function $\text{comb}_{k}$. Finally, depending on the specific task, the fused result is further transformed or updated.

\subsection{Hierarchical Physics-Informed Constraint}

Given the complexity of the AC-OPF problem, traditional GNN-based methods typically impose physics-informed constraints only on the final output, ensuring the solution remains within safe limits. However, this overlooks two critical issues in complex topologies: (1) the lack of effective regulation on dynamic feature learning hampers stability and reliability, and (2) a single constraint is insufficient for multi-layer networks, as each layer captures features at different scales and ranges. Therefore, multi-level constraints are essential to more effectively guide the learning process.

To tackle this issue, hierarchical physics-informed constraints are incorporated at the encoder stage to mitigate the deficiencies of current methods and enhance the quality of the solution. Specifically, it is essential to separately model the equality and inequality constraints in the AC-OPF problem and explicitly embed these constraints into the loss function to steer the neural network toward learning solutions that comply with physical requirements. By integrating constraint information through additional loss, the model effectively captures their influence during backpropagation, thereby enhancing the physical feasibility and optimization quality of the resulting solutions. Specifically, it is first necessary to use a GCN to extract high-dimensional features from the input graph topology data; here, we adopt a GCN with an attention mechanism~\citep{shi2020masked} as the backbone network. Given the node graph $\mathcal{G}=(V,E)$ and its corresponding node feature representation $F^{(l)} = \{ v_1^{(l)}, v_2^{(l)}, \dots, v_n^{(l)} \}$, where $v_i^{(l)} \in \mathbb{R}^d$ denotes the feature vector of node $i = 1, 2, \dots, n$, with $d$ representing the feature dimension and $n$ indicating the total number of nodes, the feature interaction between each pair of adjacent nodes $(i,j)$ is computed using a multi-head attention mechanism. This process is expressed by the following equation:

\begin{equation}
\begin{aligned}
q_{c,i}^{(l)} &= W_{c,q}^{(l)} v_i^{(l)} + b_{c,q}^{(l)} \\
k_{c,j}^{(l)} &= W_{c,k}^{(l)} v_j^{(l)} + b_{c,k}^{(l)} \\
e_{c,ij} &= W_{c,e} e_{ij} + b_{c,e} \\
\alpha_{c,ij}^{(l)} &= \frac{\exp\left(\frac{(q_{c,i}^{(l)})^T (k_{c,j}^{(l)} + e_{c,ij})}{\sqrt{d_c}}\right)}{\sum_{u \in N(i)} \exp\left(\frac{(q_{c,i}^{(l)})^T (k_{c,u}^{(l)} + e_{c,iu})}{\sqrt{d_c}}\right)}
\label{1}
\end{aligned}
\end{equation}

\noindent where $d_c$ is the hidden size of the \(c\)-th head. The features of node $v_i^{(l)}$ and $v_j^{(l)}$ are first linearly transformed into the query vector \(q\), key vector \(k\), and edge feature vector \(e\) using distinct trainable parameters $W_{c,q}^{(l)}$, $W_{c,k}^{(l)}$, and $W_{c,e}$, along with the corresponding biases $b_{c,q}^{(l)}$, $b_{c,k}^{(l)}$, and $b_{c,e}$, to compute the attention score for the $c$-th head. Subsequently, the computed multi-head attention scores are utilized to aggregate the feature information from node $i$ to node $j$:

\begin{equation}
\begin{aligned}
     h_{c,j}^{(l)} = W_{c,v}^{(l)}v_{j}^{(l)} + b_{c,v}^{(l)}
\end{aligned}
\end{equation}

\begin{equation}
\begin{aligned}
    \hat{v}_i^{(l+1)} = \left\| \sum_{c=1}^{C} \left[ \sum_{j \in N(i)} \alpha_{c,ij}^{(l)} \left (h_{c,j}^{(l)} + e_{c,ij} \right) \right] \right\|
\end{aligned}
\end{equation}

\noindent where the ‖ is the concatenation operation for $C$ head attention. The  feature $v_j$ is transformed to $\bm{h}_{c,j} \in \mathbb{R}^d$ for weighted sum. Subsequently, a gated residual connection between layers, as shown in Equation(\ref{eq:update}), is used to prevent the model from oversmoothing.

\begin{align}
r_i^{(l)} &= W_r^{(l)} v_i^{(l)} + b_r^{(l)} \notag \\
\beta_i^{(l)} &= \text{sigmoid}\left( W_g^{(l)} \left[ \hat{v}_i^{(l+1)}; \hat{r}_i^{(l+1)}; v_i^{(l)} - r_i^{(l)} \right] \right) \notag \\
v_i^{(l+1)} &= \sigma\left( \text{LayerNorm}\left( (1 - \beta_i^{(l)}) \hat{v}_i^{(l+1)} + \beta_i^{(l)} r_i^{(l)} \right) \right) \label{eq:update}
\end{align}

\noindent where $r_i$ and $\beta_i$ denote the residual feature and gating factor, respectively.

\begin{align}
\hat{v}_i^{(l+1)} &= \frac{1}{C} \sum_{c=1}^{C} \left[ \sum_{j \in N(i)} \alpha_{c,ij}^{(l)} (h_{c,j}^{(l)} + e_{c,ij}^{(l)}) \right] \label{eq:hat_h} \\
v_i^{(l+1)} &= (1 - \beta_i^{(l)}) \hat{v}_i^{(l+1)} + \beta_i^{(l)} r_i^{(l)} \label{eq:h}
\end{align}

Apply a linear transformation to the output of the $l$-th layer to obtain $P_g$, $Q_g$, $V$ and $\theta$:

\begin{equation}
\phi^l = \begin{bmatrix}
P_g \\
Q_g \\
V \\
\theta
\end{bmatrix}
= W^{(l)} v^{(l)} + b^{(l)}
\label{loss1}
\end{equation}

\noindent where \( v^{(l)} \) is the output of the \( l \)-th layer, \( W^{(l)} \in \mathbb{R}^{4 \times d} \) is the weight matrix, and \( b^{(l)} \in \mathbb{R}^4 \) is the bias vector. 

Next, Physics-Informed Neural Networks with Hard Constraints (hPINNs)~\citep{lu2021physics} are utilized to constrain the obtained $\phi$ values. Specifically, these physics-informed constraints ensure that both equality constraints (Equations (\ref{nodal_p})–(\ref{nodal_q})) and inequality constraints (Equations (\ref{ineq:generator_power_p})–(\ref{ineq:generator_power_θ})) are consistently satisfied, while maintaining the branch power at each node within its predefined limits. The equality constraints $\bm{\mathcal{E}}$ (\ref{nodal_p})-(\ref{nodal_q}), also known as nodal balance equations, are defined as follows:

\begin{equation}
P_{g,i} - P_{d,i} = \text{Re} \left[ V_i \sum_{j=1}^n Y_{ij} V_j^* \right]
\label{nodal_p}
\end{equation}

\begin{equation}
Q_{g,i} - Q_{d,i} = \text{Im} \left[ V_i \sum_{j=1}^n Y_{ij} V_j^* \right]
\label{nodal_q}
\end{equation}

\noindent Here ${P}_{g,i}$ and ${Q}_{g,i}$ denote the active and reactive power of the $i$-th generator, respectively, while ${P}_{d,i}$ and ${Q}_{d,i}$ represent the active and reactive power of the $i$-th load, respectively. \(V_i = e_i + jf_i\) represents the complex voltage at bus \(i\); \(Y_{ij} = G_{ij} + jB_{ij}\) denotes the element of the admittance matrix; \(\text{Re}[\cdot]\) and \(\text{Im}[\cdot]\) indicate the real and imaginary parts, respectively. Additionally, \(V_j^* = e_j - jf_j\) is the complex conjugate of the voltage at bus \(j\), which is the conjugate form of \(V_j = e_j + jf_j\), used for calculating complex power. The inequality constraints $\bm{\mathcal{I}}$ include limits on the generator power (\ref{ineq:generator_power_p})-(\ref{ineq:generator_power_q}), voltage magnitude (\ref{ineq:generator_power_v}), and branch power at each node (\ref{ineq:generator_power_θ}), defined as follows:

\begin{equation}
\mathcal{I}_P = \bigcup_{r \in N_g} \{ P_{g,r,\min} \leq P_g \leq P_{g,r,\max} \}
\label{ineq:generator_power_p}
\end{equation}

\begin{equation}
\mathcal{I}_Q = \bigcup_{r \in N_g} \{ Q_{g,r,\min} \leq Q_g \leq Q_{g,r,\max} \}
\label{ineq:generator_power_q}
\end{equation}

\begin{equation}
\mathcal{I}_V = \bigcup_{i \in N} \{ V_{i,\min} \leq V_i \leq V_{i,\max} \}
\label{ineq:generator_power_v}
\end{equation}

\begin{equation}
\mathcal{I}_S = \bigcup_{(ij) \in E} \{ (P_{ij})^2 + (Q_{ij})^2 \leq |S_{\max,ij}|^2 \}
\label{ineq:generator_power_θ}
\end{equation}

\noindent where \( P_{ij} \), \( Q_{ij} \), and \( S_{ij} \) represent the active power, reactive power, and apparent power transmitted between the \(i\)-th and \(j\)-th nodes (or buses), respectively.

Then, we derive the corresponding
mathematical representations of the equality and inequality constraints. \\ 

\noindent \textbf{Equality loss:}

\begin{equation}
\begin{bmatrix}
P_g^{'} \\
Q_g^{'}
\end{bmatrix}
=
\mathrm{M_{GtB}} \cdot
\begin{bmatrix}
{P}_g \\
{Q}_g
\end{bmatrix}
\end{equation}

\begin{equation}
\begin{bmatrix}
{P_g^{loss}} \\
{Q_g^{loss}}
\end{bmatrix}
=
\mathrm{(M_{EtB})}^T \cdot
\begin{bmatrix}
{P}_f^{net} \\
{Q}_f^{net}
\end{bmatrix}
\end{equation}

\begin{equation}
\mathcal{L}_{eq} = \left| \begin{bmatrix} {P_g^{loss}} \\ {Q_g^{loss}} \end{bmatrix} + \begin{bmatrix} P_d \\ Q_d \end{bmatrix} - \begin{bmatrix} P_g^{'} \\ Q_g^{'} \end{bmatrix} \right|
\end{equation}

\noindent where \(\mathrm{M_{GtB}}\) and \(\mathrm{(M_{EtB})}^T\) denote the mapping matrix from generators to their connected nodes and the transpose of the relationship mapping matrix between the ending node of each edge and all nodes in the network, respectively. \(P_f^{net}\) and \(Q_f^{net}\) denote the active and reactive power flows on each transmission line, respectively, computed using Equations (\ref{flow:pf}) and (\ref{flow:qf}). \\ 

\noindent \textbf{Inequality loss:}

\begin{equation}
\mathcal{L}_{ineq}^{P} =
\begin{bmatrix}
P_{max}^{loss} \\
P_{min}^{loss}
\end{bmatrix}
=
\begin{bmatrix}
\sigma(P_g - P_{max}) \\
\sigma(P_{min} - P_g)
\end{bmatrix}
\end{equation}

\begin{equation}
\mathcal{L}_{ineq}^{Q} =
\begin{bmatrix}
Q_{max}^{loss} \\
Q_{min}^{loss}
\end{bmatrix}
=
\begin{bmatrix}
\sigma(Q_g - Q_{max}) \\
\sigma(Q_{min} - Q_g)
\end{bmatrix}
\end{equation}

\begin{equation}
\mathcal{L}_{ineq}^{V} =
\begin{bmatrix}
V_{max}^{loss} \\
V_{min}^{loss}
\end{bmatrix}
=
\begin{bmatrix}
\sigma(V - V_{max}) \\
\sigma(V_{min} - V)
\end{bmatrix}
\end{equation}

\begin{equation}
    \mathcal{L}_{ineq} = \mathcal{L}_{ineq}^{P} + \mathcal{L}_{ineq}^{Q} + \mathcal{L}_{ineq}^{V}
\end{equation}

\noindent Here, \(P_{max}\), \( Q_{\text{max}} \), and \( V_{\text{max}} \) denote the maximum limits of active power, reactive power, and voltage magnitude, respectively, for a specified generator node, whereas \(P_{min}\), \( Q_{\text{min}} \), and \( V_{\text{min}} \) denote the corresponding minimum limits of active power, reactive power, and voltage magnitude. \( \sigma \) represents the ReLU activation function, which ensures that a valid gradient is computed only when the input exceeds the maximum limit or falls below the minimum limit. \\ 

\noindent \textbf{Power flow loss:}

\begin{equation}
    \theta_{diff} = \mathrm{M_{EfB}} \cdot {\theta} - \mathrm{M_{EtB}} \cdot {\theta} 
\end{equation}

\begin{align}
\left\{
\begin{aligned}
P_{f1}^{tmp} &= V_f \cdot V_t \cdot \left( G \cdot \cos(\theta_{diff}) - B \cdot \sin(\theta_{diff}) \right) \\
P_{f2}^{tmp} &= (V_f)^2 \cdot G \cdot t \\
P_f^{net} &= P_{f2}^{tmp} - P_{f1}^{tmp}
\end{aligned}
\right.
\label{flow:pf}
\end{align}

\begin{align}
\left\{
\begin{aligned}
Q_{f1}^{tmp} &= V_f \cdot V_t \cdot \left( B \cdot \cos(\theta_{diff}) - G \cdot \sin(\theta_{diff}) \right) \\
Q_{f2}^{tmp} &= -(V_f)^2 \cdot \left( B + \frac{sh}{2} \right) \cdot l \\
Q_f^{net} &= Q_{f2}^{tmp} + Q_{f1}^{tmp}
\end{aligned}
\right.
\label{flow:qf}
\end{align}

\begin{align}
\mathcal{L}_{flow} &= \sigma\left( (P_f^{net})^2 + (Q_f^{net})^2 - (S_{max})^2 \right) \label{eq:flow_loss_raw}
\end{align}

\noindent Here, \(\mathrm{M_{EfB}}\) denotes the relationship mapping matrix between the starting node of each edge and all nodes in the network, while \(\theta_{diff}\) represents the phase angle difference across each edge. Additionally, \(V_f=\mathrm{M_{EfB}}\cdot V\) and \(V_t=\mathrm{M_{EtB}}\cdot V\) indicate the voltage magnitudes at the starting and ending nodes of each edge, respectively. Furthermore, \(t\) signifies the transformer tap ratio. Similarly, \(G\), \(B\), \(sh\) and \(t\) denote the conductance, susceptance, shunt admittance, and transformer tap ratio of the transmission line, respectively. Lastly, \(S\) represents the maximum apparent power of the transmission line. \\ 

\noindent \textbf{Total loss:}

\begin{equation}
    \mathcal{L}^{pinn} =  \mathcal{L}_{\text{flow}} + \mathcal{L}_{\text{eq}} + \mathcal{L}_{\text{ineq}} 
\end{equation}

\begin{equation}
\mathcal{L}_{opf} = \sum_{i=1}^{\text{len}(\phi^L)} \left( \phi_i^L - \phi_i^{\text{target}} \right)^2
\end{equation}

\begin{equation}
    \mathcal{L}_{total} = \mathcal{L}_{\text{opf}} + \sum_{i}^{L} \alpha_i\mathcal{L}^{pinn}_i
\end{equation}

\noindent Here, \(\mathcal{L}_{opf}\) denotes the objective function for optimizing the AC-OPF model, while \(\phi_{i}^{target}\) represents the solution derived from the MIPS~\citep{zimmerman2016matpower}, satisfying power balance, voltage constraints, and power flow constraints, and serving as the target reference for the model. \(\mathcal{L}_{opf}\) represents the objective function for the model optimization of the AC-OPF problem, \(L\) represents the number of layers in the network. Furthermore, given that different layers of a deep learning model capture information across various spatial dimensions, we assign distinct weights \(\alpha\) to each layer while ensuring that their sum equals 1, thereby improving the robustness of the model’s predictions.

\subsection{Dynamic Domain Adaption Mechanism}

Inspired by the strength of hierarchical networks in modeling diverse long-range dependencies, we introduce the application of varying levels of physical constraints across the layers of the graph Transformer architecture. Fixed constraint strategies may hinder the network’s ability to adaptively learn knowledge representation by restricting its receptive field. Given the inherent flexibility of physical constraints, they should be dynamically adjusted according to the network architecture and task requirements rather than being applied rigidly. Unlike previous approaches that impose constraint losses exclusively at the output layer, this work proposes a more adaptive, multi-layer constraint strategy. As illustrated in Figure~\ref{pipeline_fig}, the Dynamic Domain Adaptation (DDA) mechanism is proposed to adjust constraint intensity based on the number of samples in each layer that satisfy or violate predefined constraints, enabling more flexible and effective feature learning across layers.

Specifically, given the graph-structured data $V = {v_1, v_2, \dots, v_n}$, where $n$ denotes the number of nodes in the graph, high-dimensional features are first extracted through the layers of the GCN. Then, the feedforward neural network is used to map these features to the corresponding target variables $\phi \in \mathbb{R}^{n \times 4}$, where each node corresponds to four target variable values ($P_g$, $Q_g$, $V$, and $\theta$). The predicted target variables $\phi$ together with the input features $P_{d}$ and $Q_{d}$ are evaluated against the predefined physical constraints to identify violation patterns. The following sets are defined to categorize constraint violations:

\begin{equation}
\left\{
\begin{aligned}
S_{\mathcal{I}}^{+} = \{ i \in \{1,2,\dots,n\} \mid \phi_i > \mathcal{I}_i^{\text{max}} \} \\
S_{\mathcal{I}}^{-} = \{ i \in \{1,2,\dots,n\} \mid \phi_i < \mathcal{I}_i^{\text{min}} \}
\end{aligned}
\right.
\end{equation}

\begin{equation}
\left\{
\begin{aligned}
S_{\mathcal{E}}^{+} = \{i \mid func(\phi_i,P_{d},Q_{d}) > \mathcal{E}_{i} \} \\
S_{\mathcal{E}}^{-} = \{i \mid func(\phi_i,P_{d},Q_{d}) < \mathcal{E}_{i} \}
\end{aligned}
\right.
\end{equation}

\noindent where, $func$ is the equation and $\mathcal{S}$ is the set of points that do not satisfy the constraints. Then, the violation ratio for each set of constraint violations is computed according to the number of samples, and this ratio is employed to dynamically modify the original constraint limits. The detailed adjustment formula is given below:

\begin{equation}
\left\{
\begin{aligned}
\mathcal{I}_i^{\text{max}} = \mathcal{I}_i^{\text{max}} + \mathcal{I}_i^{\text{max}}\cdot func^{*}(S_{\mathcal{I}}^{+}) \\
\mathcal{I}_i^{\text{min}} = \mathcal{I}_i^{\text{min}} - \mathcal{I}_i^{\text{min}}\cdot func^{*}(S_{\mathcal{I}}^{-})
\end{aligned}
\right.
\end{equation}

\begin{equation}
\mathcal{E}_i = \mathcal{E}_i + \mathcal{E}_i\cdot (|func^{*}(S_{\mathcal{E}}^{+})-func^{*}(S_{\mathcal{E}}^{+})|) 
\end{equation}

\noindent where the function $func^*$ is used to calculate the violation ratio of constraints, dynamically updating the constraint ranges of subsequent layers to accommodate the differing modeling capabilities at various network levels, thereby ensuring the robustness of the network throughout the knowledge acquisition process.

\subsection{Temporal Multi-Feature Extraction}

Recent studies have primarily focused on utilizing GCNs to extract structural information from power grid topologies for solving the AC-OPF problem. However, these methods typically neglect the importance of multi-factor modeling in load forecasting, particularly the incorporation of temporal dynamics. Indeed, the generator’s output is closely related to both the number of connected loads and their electrical distances—factors that should be properly considered in modeling. To address this, we propose a method that uses one-dimensional convolution to process serialized graph features, aiming to improve load forecasting by introducing distance-based serialized information. We first apply Dijkstra’s algorithm to order the graph data based on key spatial features, including \textbf{Conductance}, \textbf{Susceptance}, and \textbf{Adjacency Distance}. Then, one-dimensional convolutions are used to extract meaningful patterns from these serialized features.

\begin{algorithm}[H]
\caption{Dijkstra Algorithm}
\label{alg:algorithm}
\textbf{Input}: Graph $\mathcal{G}(V,E)$, starting node $s$, edge weights, conductance, susceptance $w(G,B,D) \geq 0$
\textbf{Output}: Shortest path distances from $s$ to all nodes
\begin{algorithmic}[1] %
\STATE Let $d[s] = 0$, $d[v] = \infty$ for all $v \neq s$.
\STATE Create priority queue $Q$ with all nodes, initialized with distances.
\STATE Create set $S$ (visited nodes), initially empty.
\WHILE{$Q$ is not empty}
\STATE Extract node $u$ with minimum distance from $Q$.
\STATE Add $u$ to $S$.
\IF{$u$ has unvisited neighbors $v$}
    \FOR{each neighbor $v$ of $u$}
        \IF{$d[u] + w(u, v) < d[v]$}
            \STATE Update $d[v] = d[u] + w(u, v)$.
            \STATE Update $v$'s priority in $Q$.
        \ENDIF
    \ENDFOR
\ENDIF
\ENDWHILE
\STATE \textbf{return} $d[v]$ for all $v$
\end{algorithmic}
\end{algorithm}

Specifically, given a node graph \( \mathcal{G} = (V, E) \) and its corresponding node feature matrix \( F \in \mathbb{R}^{|V| \times d_v} \), where \( d_v \) represents the predefined dimension of features based on \( P_d \) and \( Q_d \). On this basis, select one generator node \(v_q\) as the starting node of the sequence. Then, use Dijkstra's algorithm to sort the remaining nodes (including load nodes and other unselected generator nodes), with the sorting criteria being conductance (\(G\)), susceptance (\(B\)), and adjacency distance, respectively. These features serve as the weights for the algorithm. The above process can be expressed as the following formula:

\begin{equation}
d(v_s, v_j) = \min_{P: v_s \to v_j} \sum_{(v_i, v_j) \in P} \left( \alpha \cdot \frac{1}{G^2_{ij}} + \beta \cdot \frac{1}{B^2_{ij}} + \gamma \cdot D_{ij} \right)
\end{equation}

\noindent where \( d(v_s, v_j) \) denotes the optimal distance from the starting node \( v_s \) to the target node \( v_j \), while \(P_{v_s \sim v_j}\) represents all possible paths from \( v_s \) to \( v_j \). \( G_{ij} \) and \( B_{ij} \) signify the conductance and susceptance between nodes \( i \) to node \(j \), respectively, with \(D_{ij} \) assigned a constant value of 1. We designate resistance \(R\) (i.e., the reciprocal of conductance, measured in siemens, S) and reactance \(X\) (i.e., the reciprocal of susceptance, measured in siemens, S) as the two primary reference parameters for the algorithm. The selection of \(R\) and \(X\) is driven by their ability to optimize paths, minimizing energy loss and voltage fluctuations, thereby enhancing the precision in modeling the electrical characteristics of transmission lines and improving the accuracy of serialized features. The parameters \(\alpha\), \(\beta\), and \(\gamma\) function as weighting coefficients to regulate the contributions of various factors, ensuring a balanced influence of resistance, reactance, and adjacency distance in the ranking process. Subsequently, all \( \min(d(v_s, v_j)) \) values are arranged in ascending order to derive the ordered node sequence \( S_N = \{ v_q, v_{s1}, v_{s2}, \dots, v_{sj} \} \), where \( j \in (|V| - 1) \):

\begin{equation}
    S_N = \text{sort}_{v_j \in V \setminus \{v_q\}} (\min_{v_k \in Q_l}(d(v_s, v_j)))
\end{equation}

Next, the ordered node sequence \( S_N \) is utilized to reorganize the original unordered node features \( F \in \mathbb{R}^{|V| \times d_v} \), yielding the reordered node features \(F_{\text{sorted}} = [F_{j_1}, F_{j_2}, \ldots, F_{j_n}] \in \mathbb{R}^{|V| \times d_v}\). Then, the reordered node features are fed into a three-layer 1D convolutional neural network to extract temporal features.

\begin{equation}
H^k_{TMFA} = \sigma(\text{Conv1D}(H_{TMFA}^{k-1}, W^k, b^k)), \quad k = 1, 2, 3
\end{equation}

\noindent where \(W_l\) and \(b_l\) represent the weights and biases for layer \(l\), respectively, and \(H_{TMFA}^{0} = H^{0}\). The extracted temporal features are added as residuals to the original features to enhance feature representation. Subsequently, the extracted features with temporal characteristics are incorporated into the original features in a residual form, and then transformed into high-dimensional features through linear transformation.

\begin{equation}
    H^{0,*} = H^{0} + H^{3}_{TMFA}
\end{equation}

\noindent where \(H^{0,*}\) are used as the input for the subsequent layers of the GCN network.

\section{Experiments}
\subsection{Datasets and Evaluation Protocol}

Following established research methods~\citep{huang2021deepopf, varbella2024physics, nellikkath2022physics}, we perform comprehensive experiments across multiple power networks, including the IEEE 9, 14, 30, 39, 57, 118, 145, and 300-bus systems, employing an 80-20\% training-test split to assess the effectiveness of the proposed method. The IEEE power system test cases represent grids with varied characteristics and serve as a standard benchmark for evaluating AC-OPF approaches. Detailed specifications for each test case are provided in Table~\ref{test cases}. All test cases are developed using the MATPOWER package~\citep{zimmerman2020matpower}, which leverages the MATPOWER Interior node Solver (MIPS)~\citep{zimmerman2016matpower}, a widely recognized and utilized solver in power system analysis. Specifically, the load values for each load bus are drawn from a uniform distribution, with a variation range of ±10\% around the default load, using randomly sampled load scenarios. Subsequently, MIPS is employed to solve the sampled loads, and the resulting optimal solution is treated as the ground truth for the corresponding load.

\begin{table}[H]
    \fontsize{8pt}{10pt}\selectfont
    \centering
    
    \begin{tabular}{ccccccc}
        \toprule
        Power Grid & $N_b$ & $N_d$ & $N_g$ & $N_l$ & MW & MVA\\
        \midrule
        IEEE-9 & 9 & 9 & 3 & 13 & 315 & 115 \\
        IEEE-14 & 14 & 11 & 5 & 20 & 259 & 200\\
        IEEE-30 & 30 & 21 & 6 & 41 & 283 & 126\\
        IEEE-39 & 39 & 21 & 10 & 46 & 6254 & 6626 \\
        IEEE-57 & 57 & 42 & 7 & 80 & 1250 & 327\\
        IEEE-118 & 118 & 99 & 19 & 186 & 4242 & 4537\\
        IEEE-145 & 145 & 102 & 50 & 401 & 18300 & 19000 \\
        IEEE-300 & 300 & 200 & 69 & 411 & 23530 & 12100\\
        \bottomrule
    \end{tabular}

\caption{Test Case Characteristics.}
\label{test cases}
\end{table}

We use the following metrics to evaluate the performance of DDA-PIGCN:

\subsubsection{Mean Absolute Error.}It quantifies the Mean Absolute Error (MAE) between the true values obtained from the MIPS solver and the predicted values generated by DDA-PIGCN.

\subsubsection{Constraint Satisfaction.}As a key metric for assessing the effectiveness of physics-informed constraints, it evaluates the feasibility of the proposed approach from two perspectives: the constraint satisfaction ratio (i.e., the percentage of inequality constraints satisfied) and the degree of violation (i.e., the distance between the violated variable and the boundary). Use \(\kappa{P_g}\)(\(\delta{P_g}\)), \(\kappa{Q_g}\)(\(\delta{Q_g}\)), \(\kappa{S_l}\), and \(\kappa\theta_l\) to represent the constraint satisfaction ratios (the degree of violation) ratios of active generation, reactive generation, branch power flow, and phase angle difference, respectively.

\subsubsection{Probabilistic Accuracy.}It measures the accuracy of the predicted values within a defined range.
\subsection{Implementation Details}

For DDA-PIGNN, we adopt UniMP~\citep{shi2020masked} as the backbone, following the setup in PINCO~\citep{varbella2024physics}. The architecture includes 8 Graph Transformer layers, followed by two linear layers with Tanhshrink as the activation function. Each intermediate layer has a hidden dimension of 24 and uses 4 multi-head attention mechanisms. Models are implemented and trained using PyTorch, leveraging its flexibility and rich ecosystem. Training is performed on an NVIDIA GeForce RTX A100 (40G) GPU, which provides efficient processing for large models and datasets. An exponential learning rate decay is applied with a decay factor $\gamma$ of 0.996. Hyperparameters related to the H-PINN method (e.g., $\mu_{g}$, $\mu_{H}$, $\beta_{G}$, $\beta_{H}$) follow values recommended in prior work. Table~\ref{2222} lists additional settings such as batch size and initial learning rate.

\begin{table}[H]
    \fontsize{8pt}{10pt}\selectfont
    \centering
    
    \begin{tabular}{cccccc}
        \toprule
        Test Case & Number & Batch & Learning Rate & Epoch & Gamma \\
        \midrule
        IEEE-9 & 30000 & 256 & 1e-5 & 10000  & 0.9995\\
        IEEE-14 & 30000 & 256 & 1e-5 & 10000 & 0.9995\\
        IEEE-30 & 10000 & 128 & 3e-5 & 10000 & 0.9995\\
        IEEE-39 & 10000 & 128 & 3e-4 & 10000  & 0.9995\\
        IEEE-57 & 8000 & 128 & 5e-3 & 15000 & 0.9995\\
        IEEE-118 & 5000 & 64 & 4e-3 & 15000 & 0.9995\\
        IEEE-145 & 5000 & 64 & 7e-4 & 20000 & 0.9995\\
        IEEE-300 & 2000 & 64 & 7e-4 & 20000 & 0.9995\\
        \bottomrule
    \end{tabular}

\caption{Hyperparameter Settings}
\label{2222}
\end{table}

\subsection {Main Results}

As presented in Table~\ref{table 3}, we assess the feasibility of DDA-PIGCN in addressing the AC-OPF problem within power grid systems by evaluating constraint satisfaction and the degree of constraint violation. Here, M1 and M2 denote the OPF-HGNN~\citep{ghamizi2024opf} and Deepopf-V~\citep{huang2021deepopf} methods, respectively. The results demonstrate that DDA-PIGCN achieves better constraint feasibility than the baselines, more effectively keeping variable states within predefined bounds and thus improving both physical feasibility and optimization performance. Moreover, evaluations on diverse test cases, including large-scale systems like IEEE-118, confirm the model’s strong generalization ability, stability, and superior constraint handling in complex grid environments.

\begin{table}[H]
    \fontsize{8pt}{10pt}\selectfont
    \centering
    
    \begin{tabular}{l|ccc|ccc}
        \toprule
        \multirow{2}{*}{Metric} & \multicolumn{3}{c|}{IEEE 39-Bus System} & \multicolumn{3}{c}{IEEE 118-Bus System} \\
        \cmidrule(lr){2-4} \cmidrule(lr){5-7}
        & \textbf{Ours}  &\(M_1\)  & \(M_2\) & \textbf{Ours}  & \(M_1\)  & \(M_2\)   \\
        \midrule
        \(\kappa{P_g}\) & 100.0 & 98.9 & 100.0 & 100.0  & 96.2 & 100.0 \\
        \(\kappa{Q_g}\)   & 100.0 & 99.9 & 99.7 & 100.0  & 95.8 & 100.0 \\
        \(\kappa{V}\)   & 99.6 & 97.2 & 100.0 & 100.0  & 100.0 & 99.8 \\
        \(\kappa{S_l}\)   & 100.0 & 100.0 & 100.0 & 100.0  & 100.0 & 100.0  \\
        \(\kappa{\theta_l}\)   & 100.0 & 100.0 & 100.0 & 100.0  & 95.27 & 100.0  \\
        \( \delta{P_g} \)   & 0.0022 & 0.0023 & 0.0027 & 0.0102  & 0.0221 & 0.0155 \\
        \( \delta{Q_g} \)  & 0.0029 & 0.0052 & 0.0036 & 0.0072  & 0.0157 & 0.0121 \\
    
        \bottomrule
    \end{tabular}
    \caption{The comparison of constraint feasibility between \textbf{DDA-PIGCN} and other methods is presented across various test cases.}
    \label{table 3}
\end{table}

\begin{table*}[t]
    \fontsize{8pt}{10pt}\selectfont
    \centering
    
    \begin{tabular}{l|cccc|cccc|cccc}
        \toprule
        \multirow{2}{*}{Test Case} & \multicolumn{4}{c|}{\textbf{DDA-PIGCN (Ours)}} & \multicolumn{4}{c|}{OPF-HGNN~\citep{ghamizi2024opf}} & \multicolumn{4}{c}{Deepopf-V~\citep{huang2021deepopf}} \\
        \cmidrule(lr){2-5} \cmidrule(lr){6-9} \cmidrule(lr){10-13}
        & \( P_g \)  & \( Q_g \)  & \( V \) & \( \theta \)  & \( P_g \) & \( Q_g \) & \( V \) & \( \theta \) & \( P_g \) & \( Q_g \) & \( V \) & \( \theta \) \\
        \midrule
        IEEE 9-Bus System   & 0.0011 & 0.0017 & 0.0010 & 0.0007  & 0.0016 & 0.0036 & 0.0012  & 0.0017  & 0.0101 & 0.0078 & 0.0082 &0.0129  \\
        IEEE 14-Bus System  & 0.0023 & 0.0113 & 0.0001 & 0.0014  & 0.0032 & 0.0113 & 0.0043  & 0.0087  & 0.0077 & 0.0104 & 0.0218 &0.0191  \\
        IEEE 30-Bus System  & 0.0042  & 0.0133 & 0.0029 & 0.0073  & 0.0025 & 0.0321 & 0.0129  & 0.0311  & 0.0382 & 0.0157 & 0.0410 &0.0404 \\
        IEEE 39-Bus System  & 0.0073 & 0.0130 & 0.0021 & 0.0039  & 0.0212 & 0.0156 & 0.0053  & 0.0714  & 0.0409 & 0.0451 & 0.0271 &0.0328  \\
        IEEE 57-Bus System  & 0.0024 & 0.0031 & 0.0012 & 0.0023  & 0.0281 & 0.0417 & 0.0288  & 0.0301  & 0.0602 & 0.0584 & 0.0587 &0.0492  \\
        IEEE 118-Bus System & 0.0015 & 0.0074 & 0.0024 & 0.0312  & 0.0563 & 0.0764 & 0.0557  & 0.0666  & 0.0613 & 0.0569 & 0.0727 &0.1061  \\
        IEEE 300-Bus System & 0.0514 & 0.0721 & 0.0332 & 0.0928  & 0.0754 & 0.0718 & 0.0975  & 0.1084  & 0.1782 & 0.1424 & 0.0942 &0.0998  \\
    
        \bottomrule
    \end{tabular}
    \caption{Quantitative results for the Mean Absolute Error (MAE) between the predicted and ground truth values.} 
    \label{table 2}
\end{table*}

Furthermore, to evaluate the potential of DDA-PIGCN in solving for the optimal solution, we calculate the MAE between the predicted and ground truth values for four variables: active power \(Pg\), reactive power \(Qg\), voltage magnitude \(V\), and phase angle \(\theta\). The results are shown in Table~\ref{table 2}. While all methods demonstrate strong performance, DDA-PIGCN achieves lower errors across multiple metrics by leveraging temporal priors embedded in the graph structure, highlighting its superior modeling accuracy and alignment with physical system behavior. Additionally, by integrating physical constraints across multiple network layers, DDA-PIGCN effectively guides the solver toward high-quality solutions under varying operating conditions. On small-scale systems such as IEEE-9 and IEEE-14, its results closely align with those of the MATPOWER solver. For larger systems like IEEE-118, despite minor prediction errors, DDA-PIGCN maintains compliance with operational requirements, demonstrating strong practicality and cross-scale generalization.

\subsection{Ablation Studies}

\begin{figure}[H]
    \centering
    \includegraphics[width=\columnwidth]{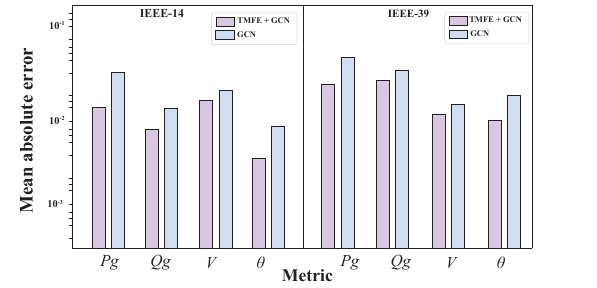}  
    \caption{The MAE results for GCN and GCN enhanced with the Temporal Multi-Feature Extraction (TMFE) module are presented for the IEEE-14 and IEEE-39 test cases.}
    \label{Figure 3}
\end{figure}

\textbf{The portability of TMFE.} The Temporal Multi-Feature Extraction (TMFE) module is a key component of DDA-PIGCN, and validating its portability is essential for future research. To evaluate this, TMFE is integrated into a standard GCN, and its performance was compared to a baseline in terms of constraint satisfaction, violation degree, and accuracy, as shown in Figure~\ref{Figure 3}. Results show that adding TMFE significantly improves performance across all metrics, suggesting that time-series features are highly relevant to the AC-OPF problem. Effectively utilizing these features provides valuable prior information and enables richer feature representations, helping to uncover hidden patterns and enhance solver efficiency.

\textbf{The Effectiveness of Model Components.} We perform ablation analysis experiments on the IEEE-9, IEEE-39, and IEEE-118 power systems to assess the individual components of DDA-PIGCN, with the findings detailed in Table~\ref{tab:component_parameters}. To ensure the reliability of the experimental outcomes, all experiments are conducted with consistent hyperparameter settings. Experimental results demonstrate that the hierarchical physics-informed constraint (HPIC) method effectively mitigates the limitations of single-constraint networks in feature modeling. Leveraging the synergistic effect of the dynamic domain adaptation (DDA) mechanism, the overall performance of the model is further enhanced. Additionally, the temporal multi-feature extraction (TMFE) module improves the model's representational capacity and solution accuracy for the AC-OPF problem by integrating temporal information.

\begin{table}[h]
    \fontsize{6.5pt}{10pt}\selectfont
    \centering
    \begin{tabular}{lccc|cccc}
    \toprule
    \multicolumn{1}{c}{\textbf{Test Case}}& \multicolumn{1}{c}{\textbf{HPIC}} & \multicolumn{1}{c}{\textbf{DDA}} & \multicolumn{1}{c}{\textbf{TMFE}} & \multicolumn{4}{|c}{\textbf{Metric}} \\
    \cmidrule(lr){5-8}
     &  &  &  & \( P_g \) & \( Q_g \) & \( V \) & \( \theta \) \\
    \midrule
     & \checkmark &  &  & 0.0036 & 0.0101 & 0.0159 &  0.0009\\ 
    IEEE-9 & \checkmark & \checkmark &  & 0.0022 & 0.0024 & 0.0020 &  0.0011\\
     & \checkmark & \checkmark & \checkmark & \textbf{0.0011} & \textbf{0.0017} & \textbf{0.0010} &  \textbf{0.0007}\\
    \midrule
     & \checkmark &  &  & 0.0219 & 0.0186 & 0.0153 &  0.0174 \\
    IEEE-39 & \checkmark & \checkmark &  & 0.0163 & 0.0188 & \textbf{0.0020} &  0.0125 \\
     & \checkmark & \checkmark & \checkmark & \textbf{0.0073} & \textbf{0.0130} & 0.0021 &  \textbf{0.0039}\\
    \midrule
     & \checkmark &  &  & 0.0131 & 0.0282 & 0.0072 &  0.0322\\
    IEEE-118 & \checkmark & \checkmark &  & 0.0083 & 0.0298 & 0.0604 &  0.0463\\
     & \checkmark & \checkmark & \checkmark & \textbf{0.0015} & \textbf{0.0074} & \textbf{0.0024} &  \textbf{0.0312}\\
    \toprule     
    \end{tabular}
\caption{Performance enhancements attributed to individual model components are evaluated using the MAE metric.}
\label{tab:component_parameters}
\end{table}

\section{Conclusion}

In this work, we propose DDA-PIGCN, a novel framework designed to integrate dynamic domain adaptation-driven hierarchical physics-informed constraint mechanism with temporal multi-feature extraction module. It systematically embeds physical constraints at different stages of the prediction process, guiding variables with distinct long-range dependencies to achieve more reasonable feature mapping. To address the temporal dynamics of the AC-OPF problem, the 1D convolution-based sequential graph feature extraction strategy is further introduced. Experimental results demonstrate that this multi-dimensional optimization strategy not only significantly improves prediction accuracy but also offers new insights for dynamic modeling and analysis of power systems, showing promising performance and application potential.

\end{document}